# How GPT-3 responds to different publics on climate change and Black Lives Matter: A critical appraisal of equity in conversational AI

Kaiping Chen, Anqi Shao, Jirayu Burapacheep, Yixuan Li

University of Wisconsin-Madison

*Kaiping Chen (corresponding author)
**Email:** kchen67@wisc.edu

## Abstract

Autoregressive language models, which use deep learning to produce human-like texts, have become increasingly widespread. Such models are powering popular virtual assistants in areas like smart health, finance, and autonomous driving, and facilitating the production of creative writing in domains from the entertainment industry to science communities. While the parameters of these large language models are improving, concerns persist that these models might not work equally for different subgroups in society. Despite growing discussions of AI fairness across disciplines, systemic metrics lack to assess what equity means in dialogue systems and how to engage different populations in the assessment loop. Grounded in theories of deliberative democracy and science and technology studies, this paper proposes an analytical framework for evaluating equity in human-AI dialogues. Using this framework, we conducted an auditing study to examine how GPT-3 responded to different subpopulations on crucial science and social issues: climate change and the Black Lives Matter (BLM) movement. Our corpus consists of over 20,000 rounds of dialogues between GPT-3 and 3290 individuals who vary in gender, race and ethnicity, education level, English as a first language, and opinions toward the issues. We found a substantively worse user experience with GPT-3 among the opinion and the education minority subpopulations; however, these two groups achieved the largest knowledge gain, changing attitudes toward supporting BLM and climate change efforts after the chat. We traced these user experience divides to conversational differences and found that GPT-3 used more negative expressions when it responded to the education and opinion minority groups, compared to its responses to the majority groups. To what extent GPT-3 uses justification when responding to minority groups is contingent on the issue. We discuss the implications of our findings for a deliberative conversational AI system that centralizes diversity, equity, and inclusion.

## Significance Statement

This paper proposes an analytical framework for evaluating equity in human-AI dialogues. We conducted an algorithm auditing study to examine how GPT-3 responded to different subpopulations on crucial science and social issues: climate change and the Black Lives Matter (BLM) movement. We found a substantively worse user experience with GPT-3 among the opinion and the education minority subpopulations; however, these two groups achieved the largest knowledge gain, changing attitudes toward supporting BLM and climate change efforts after the chat. We traced these user experience divides to conversational differences and found that GPT-3 used more negative expressions when it responded to the education and opinion minority groups, compared to its responses to the majority groups.

## Introduction

Intelligent assistants have become an inseparable part of our daily lives in recent years, from chatbots used in financial services and smart healthcare to conversational AI systems such as Alexa, Siri, and Google Assistant. These smart systems not only address simple information-seeking questions from users (1) but also assist in high-stake decision-making scenarios such as surgery, collision prevention, and criminal justice, or serve as educational tools in health persuasion and pro-social behavior (2-4).

Along with its rapid applications, scholars and practitioners have questioned to what extent these intelligent systems are built in consideration of diversity, equity, and inclusion (DEI) (5, 6). Recent scholarship has shown that these emerging machine-learning systems perform poorly in terms of accurately identifying speeches from racial minorities (7). Furthermore, anecdotal evidence has burgeoned with concerns that AI chatbots may produce sentient and conspiracy responses (8).

While AI fairness has become a heated interdisciplinary discussion in recent years, the root of the problem—inequality in conversations—has been a challenge throughout human history. This problem goes back to Ancient Athens when philosophers such as Plato and Aristotle argued that democratic dialogues are key to achieving equity in society, yet they excluded certain populations (such as women, foreigners, and enslaved people) from these dialogues (9, 10). The exclusion of voices from certain social groups in communication systems, persists today, whether in politics (11), social issues (12), or social media (13). As scholars of democracy emphasize, inclusive dialogues should not only consist of participants from diverse backgrounds and life experiences but also should be a process where people can express, listen to, and respond to diverse viewpoints (14). Yet, inequality is "always in the room" because different languages carry different socioeconomic and cultural powers, and those whose language styles signify power often dominate the conversation (15).

Powered by large language models (LLMs), conversational AI is a form of communication system. Like communications between humans, it faces the challenge of ensuring equity and inclusion in dialogues. What distinguishes conversational AI (between humans and intelligent agents) from other traditional communication systems (e.g., interpersonal communication, mass media) is its powerful and profound implication for every sector of people's lives. Yet, how these new communication AI systems work (e.g., algorithms, training dataset) is a black box to researchers due to industrial proprietorship (16). As warned by science and technology studies (STS) scholars, emerging technologies can amplify existing social disparities (wealth, power) and discrimination when such technologies are designed and deployed without input from different publics engaged in a democratic manner (17, 18). However, there is no systematic body of work to examine how technological innovations such as conversational AI might reflect and reinforce these existing social disparities, nor are there principles to articulate the rules of the relationship between technologies and humans (17). *Understanding how emerging communication technologies like conversational AI empower or disempower different social groups can bring both theoretical significance in extending theories of (deliberative) democracy to new media technologies and practical implications for how to design and deploy these technologies to benefit all social members equally.*

Drawing on theories of deliberative democracy and science and technology communication, this paper proposes an analytical framework (**Figure 1**) for assessing equity in conversational AI. The first criterion is *engaging a diverse group of users* in the assessment loop. A growing body of scholarship in computer sciences has started to audit conversational AI systems (19). Most empirics on LLMs have investigated equity in terms of gender, race, and ethnicity. Nevertheless, extensive research in social science has shown inequality in conversations also affects other social groups such as those with limited language skills and education (15), and those who hold minority viewpoints on an issue (20). Therefore, when assessing equity in conversation AI systems, it is crucial to examine beyond gender, race, and ethnicity — the often-stressed attributes in AI fairness scholarship — to investigate how populations with limited education and language skills, and who hold minority views on an issue engage with these intelligent agents compared to the majority populations in these demographic and attitudinal attributes.

The second assessment dimension examines *equity through user experiences and learning*— to what extent there is a gap between different users regarding how they feel about their chat experience with a conversational AI system, and to what extent their knowledge about an issue grows post-chat. User experience is a key criterion adopted by human-computer interaction (HCI) designers and researchers. Scholars have proposed user experience measures such as users' satisfaction, intention to use the system again, and intention to recommend the system to others (21, 22). Necessary for equity is understanding whether people's experiences of engaging with these systems vary based on their demographic and attitudinal attributes. Learning is another key criterion. Computer-mediated communication scholarship has examined how much knowledge users gain after they engage with a chatbot system about a topic area. For example, researchers find that chatbots designed with human-like persuasive and mental strategies are more likely to persuade users to increase their knowledge of healthy and pro-social behaviors (23). An equitable conversational AI system should bring comparable user experiences to different populations and nurture social learning.

The third criterion is to investigate *equity in conversations* — to what extent there is a gap in the discussion styles and sentiments when a conversational AI system responds to different users. Many empirical works have examined bias in speech recognition systems (e.g., (7, 24)rather than dialogue systems (i.e., chatbots). Dialogue systems are more complicated than speech recognition because they require these large language models to both accurately understand user input and provide relevant responses to engage in conversations with users that involve more interaction and intelligence to address users' needs. However, few studies have explored how these LLMs respond to different social groups when they discuss crucial social and science issues. Studying conversations beyond speech recognition is vital because of the prevalent use of Alexa/Siri systems in the wild. A democratic communication system includes a diverse set of language styles: the use of greeting languages, the use of justification (25) when expressing an opinion (justification includes citing facts as well as personal stories), and the use of rhetoric (e.g., appeals to emotions). Applying to human-AI communication, a conversational AI system that embodies conversational equity should respond to different social groups comparably with greetings, the use of justification in expressing opinions, and emotions.

To evaluate equity in existing conversational AI systems along these dimensions, we conducted an algorithm auditing to collect a large-scale conversational dataset between one of the most advanced conversational AI systems — GPT-3 — and online crowd workers. OpenAI's GPT-3 is an autoregressive language model family that is capable of human-like text completion tasks. Our algorithm auditing on GPT-3 provides empirical evidence to address three research questions to evaluate the extent of equity in a conversational AI system.

**RQ1.** *How do users' experience and learning outcome differ among different English-speaking subpopulations across the world when they have dialogues with GPT-3 about crucial science and social issues (i.e., climate change and BLM)?*

**RQ2.** *How does GPT-3 converse with different subpopulations on crucial science and social issues?*

**RQ3.** *How are conversational differences correlated with users' experience and learning outcome with GPT-3 on crucial science and social issues?*

Our algorithm auditing study focuses on two topics that we let our participants converse with the GPT-3 chatbot: climate change and Black Lives Matter (BLM). The first topic represents a classic controversial science issue. Research has shown how public perceptions toward it vary across populations, especially when there is a constant minority group that holds doubtful and denial attitudes that climate change is real and human-induced, and that group is difficult to persuade to think otherwise (26). Studying the topic of climate change provides an excellent opportunity to examine not only how GPT-3 responds to this group compared to the opinion majority, but also whether there might be social learning and attitudinal changes post-chat. The second topic represents a heated social issue, which has raised continuous attention from the media and public in the recent decade (27).

By operationalizing our theoretical framework, this paper is, to our knowledge, *the first study to audit how GPT-3 has conversations with different subpopulations on crucial science and social issues*. We built a User Interface for crowd workers to directly converse with GPT-3 model. The interface integrates GPT-3 API, which gives real-time responses to our participants. We ensured diversity in our participant recruitment by using screening questions and pilot testing to understand the demographic distribution of the online crowdsourcing platform.

## Results

### *A Substantive User Experience Gap in Engaging with GPT-3 for the Opinion and the Education Minority Groups*

For RQ1, which investigates how users' experiences differ after conversing with GPT-3 about crucial science and social issues, we found a substantive divide in user experiences among the opinion and the education minority groups. There is not a significant user experience divide between the race and ethnicity minority *vs*. the majority, nor between male *vs.* female participants.

**Figure 2** presents the results of OLS regressions between our major variables of interest (i.e., populations that consist of the majority vs. minority for each demographic and opinion attribute) and their user experience with GPT-3 for the two issues respectively.

After having dialogues with GPT-3 on climate change (Panel A), opinion minorities reported lower ratings ($\beta = -0.32$, $p < 0.001$) and lower satisfaction ($\beta = -0.42$, $p < 0.001$) toward GPT-3 compared to opinion majorities. Opinion minorities also reported less intention to continue the chat ($\beta = -0.61$, $p < 0.001$) and less willingness to recommend the chatbot to others ($\beta = -0.69$, $p < 0.001$). Education minorities, compared to the education majority group, also reported more negative experiences. They reported lower ratings, worse learning experiences, and less intention to continue the chat.

For user experiences after having dialogues on BLM (Panel B), opinion minorities (i.e., those who are less supportive of BLM) again reported much worse user experience with GPT-3 compared to the opinion majorities. The effect size of the user experience gap for opinion minorities in BLM is even larger compared to that for the opinion minorities in the climate change discussions.

### *Knowledge Gains are Significant for the Opinion and the Education Minority Groups*

Although the opinion and education minority groups reported much worse user experiences with GPT-3 on both issues, their attitudes toward climate change and BLM significantly changed in a positive direction post-chat (RQ1). The top left plots in Panel A and Panel B in Figure 2 present the OLS regression where the dependent variable is knowledge gain, measured by the difference between participants' post-chat and pre-chat attitudes toward an issue.

For the education minorities asked to discuss climate change, their post-chat knowledge gain is 0.07 points higher compared to the education majority groups (on a 1-5 scale). The results suggest that GPT-3 has a much larger educational value for education minority populations. For the BLM issue, there is no significant difference between the education minority *vs.* majority group in terms of their knowledge gains on the topic.

On both issues, the opinion minority groups became more supportive of both issues after the chat, and their knowledge gains are 0.2 points and 0.12 points higher than the opinion majority groups for each issue respectively. Considering that we measured participants' attitude toward each issue on a 1-5 scale, a 0.2 points difference between the majority *vs.* minority group is substantive because it accounts for nearly 6% more knowledge gains for the opinion minority groups.

To understand how this increased support among minority groups for climate change and BLM post-chat happened, we conducted issue stance analyses on all responses given by GPT-3 and our participants throughout each round of conversations. **Figure 3** shows the change in issue stance for GPT-3 (blue lines) and participants (orange lines). The percentage of participants' responses expressing support for these issues increased from 20% to 25% over the course of the conversations. There was a trend toward convergence in the issue stance between GPT-3 and participants.

*The Black Box of How GPT-3 Responds to Different Sub-Populations: A Close Look at its Deliberation Styles and Sentiment*

Different from most literature that revealed gender and race inequality in intelligent systems, we found that the conversational differences in how GPT-3 responds to different populations lie more in the *issue opinion* and the *education level* of our participants. In the dialogues on climate change (**Figure 4**, Panel A), we found that GPT-3 was more likely to cite external links in its responses (topic 10) to the opinion minority group than the opinion majority group. Further, GPT-3 was more likely to include justification and scientific research in relation to human-caused climate change in its responses (topic 9) to the education minority compared to the education majority group. Below is an example of a typical GPT-3 response that referred participants to scientific evidence:

*GPT-3: "I think it is a real phenomenon. Scientists have studied it closely and they have discovered the facts of climate change and its impacts on the Earth."*

However, for the BLM discussions (Figure 4, Panel B), we found that GPT-3 was more likely to use preference-based responses without providing justifications (i.e., topic 5, topic 7 in Panel B) when they responded to the education and the opinion minorities. Below is an example of typical GPT-3 response to our participants that is preference-based.

*GPT-3: "I do not think it would be a good idea to talk about this. As much as I do like to help you, this is a matter we truly disagree on."*

Besides examining differences in the deliberation styles of how GPT-3 responded to different subpopulations, we investigated the sentiments of its responses. Overall, we found that GPT-3 used fewer positive sentiments while having conversations with the opinion and the education minority group on both issues. In the climate change discussions (Figure 4, Panel A), GPT-3 used more words in its responses, but fewer positive sentiments to the education minorities, compared to its responses to the education majority group. For BLM discussions (Figure 4, Panel B), GPT-3 also used more words and yet more negative sentiment in its responses to the opinion minorities compared to its responses to the opinion majority group.

*Associations between GPT-3's Conversational Styles and User Experience*

Our final research question (RQ3) aims to investigate the associations between GPT-3's conversational styles and user experience and learning. We found that in climate change conversations, user experiences were positively associated with GPT-3's use of certain language features such as the number of words and positive emotions (**Figure 5**). Conversely, GPT-3's use of negative words was associated with worse participants' learning experience, lower intention to continue the conversation, and less likelihood to recommend the chatbot. Similar patterns were also observed in conversations about the BLM issue.

## Discussion

Despite rising awareness of measuring and improving fairness in AI models across disciplines in recent years (5, 6, 18, 28), there is still limited understanding of how to assess equity in conversational AI — what a democratic conversation may look like when we extend communications between humans to intelligent agents. This paper offers a starting point to propose an analytical framework for evaluating equity in conversational AI, drawing from theories in deliberative democracy, STS, and science communication. This framework first highlights that the human(s) in the assessment loop should be considered beyond the often-studied demographic traits such as gender, and race and ethnicity by expanding to consider education level, language skills, as well as their value proposition of an issue. This framework unpacks equity, not only in terms of user experience and learning across different populations with the conversational AI system but also in response styles (e.g., deliberation, sentiments) given by the conversational AI system to different populations. As one of the first studies to audit GPT-3's dialogues with different populations on crucial social and scientific issues, our findings can inform HCI designs in several ways regarding how to centralize DEI in science and technology innovation.

First, our findings about the opposing forces between user experience and knowledge gains for certain minority populations reveal a potential dilemma facing conversational AI designs: while the opinion and the education minority groups reported much worse user experience with GPT-3 compared to the opinion and the education majority groups, these two groups also achieved the largest knowledge gains, changing attitudes toward supporting human-induced climate change and BLM after the chat. This dilemma between uncomfortable user experience and positive education values of human-AI conversations echoes classic communication theories of persuasion effects. Researchers have found that certain communication strategies (e.g., using the loss frames, and fear appeals) make participants uncomfortable, yet they can be effective in persuading pro-social behavior and attitudes (29). Successful persuasion has one or three effects on public attitudes: reinforcing beliefs, revising beliefs, and increasing knowledge. Across our findings, we found all these three impacts. The opinion majority groups' attitudes are reinforced toward being more supportive of climate change and BLM after the chat. Among our opinion minority participants, their attitudes shifted towards more supportive of both issues. This could be due to their experiencing cognitive dissonance — a phenomenon when people experience discomfort when they encounter opinions that are different from theirs (30). Cognitive dissonance sometimes can fuel people to update their beliefs (31). The education minority groups became more informed after the chat and more supportive of human-induced climate change and BLM movements.

This trade-off between dissatisfactory user experience and positive knowledge gain motivates AI engineers and regulators to consider the mission(s) of a conversational AI system in terms of how to balance between them. Towards that end, our open-ended post-survey questions, which asked what users had hoped to hear from GPT-3 on the discussed topic, provides some potential user-centered solutions. One key suggestion our participants offered for enhancing their GPT-3 user experience is to avoid repetitive answers to make responses less boring and richer in vocabulary. Other participants shared that they were disappointed when GPT-3 responded that it is not human, or it cannot understand what a human asked in the chat. Exploring how to vary the language styles of LLMs to achieve this balance is vital for conversational AI designers and engineers, as our participants expressed that they would lose trust in the AI system and would not use the system again if they are not satisfied with their user experience and chat.

Second, examining AI equity beyond race, ethnicity, and gender will inform the design of conversational AI systems for more complicated tasks beyond simple Q&A exchanges. With more chatbots entering the area of social-oriented tasks, critical for enhancing diversity in LLM training datasets is understanding how an LLM may respond to subpopulations who hold various opinions (i.e., value systems about social issues) and how these opinion groups perceive the benefits and limitations of conversational AI systems. For instance, we found that while GPT-3 used more justification such as citing external links and scientific research when responding to the education and the opinion minorities on climate change, it also used more negative sentiment words in the responses, which might explain worse user experience among these subpopulations. Differently, for BLM, we observed that GPT-3 used less justification when it responded to the education and the opinion minorities. These nuances in how these intelligent agent systems respond to different publics with varying deliberative styles and sentiments, and how their response styles vary on the issue offer vital implications for studying equity in dialogue systems — any single conversational aspect such as sentiment is inadequate to understand potential biases in dialogue systems; instead, we need a theory-driven approach, e.g., theories of deliberative democracy, to guide multi-dimensional metrics to investigate different aspects of dialogues that can contradict.

The results of GPT-3's less use of justification and more negative sentiment toward the opinion and the education minority groups on the BLM issue can bring unintended consequences for these groups. As extensive literature in human-to-human communication notes, during public deliberation about controversial social issues, participants who hold minority opinions can spiral into silence when they perceive their opinions are the minority in society; they are less likely to speak during the discussion, and thus their voices are under-heard (20). In a word, human-AI conversations often mirror inequalities in human communications. Breaking from the persistent challenge facing humanity requires more listening to these minority voices throughout the AI system design process.

Our paper offers the starting point for a critical investigation of what equity means in conversational AI and the status of equity and inclusion in LLM. To move forward, future research can expand the scope of the issues for auditing dialogues. Moreover, although collecting dialogues from existing chatbots is a challenge due to industry proprietary, it is still valuable to examine other LLMs and verify how our findings hold across different conversational AI systems. Finally, in our auditing study, participants typed their answers, which did not allow us to examine other important demographic traits such as the role of accents and tones in dialogue systems. User experiences from minority groups can differ when they use written vs. oral communication. These are all exciting areas for a more deliberative conversational AI system.

## Materials and Methods

We briefly describe our algorithm auditing design for data collection, measurements for user experience, methods for analyzing human-AI dialogues, as well as our statistical models to address the three RQs. Further details are provided in *Supplemental Information (SI)*.

### *Data and Auditing Design*

We adopted Generative Pre-trained Transformer 3 (GPT-3) from OpenAI as the language generator for our chatbot. OpenAI's GPT-3 is an autoregressive language model family that is capable of human-like text completion tasks and these tasks can be tweaked to generate conversations. GPT-3 achieved its best performance due to its large parameter capacity of 175 billion parameters. Although it is not the current state-of-the-art model, the model's architecture is based on the transformer architecture which is being used extensively in language models in the past couple of years. We have seen many successors such as LaMDA and MUM, which are based on the transformer as well. In terms of implementation, OpenAI provides us with text completion API, which we utilized in a chatbot manner to collect the dialogue dataset our participants had with GPT-3. For the GPT-3 models, we used the most capable version available at the time of data collection (text-davinci-001), and we provided details in SI Appendix 1.3 about how we used content filter configuration and the code to replicate our web application. We performed a series of tests to validate the consistency of the responses generated by GPT-3 such as whether it can recognize and give similar output on synonyms and double negatives (See SI Appendix A1.4).

Our auditing design follows three stages: a pre-dialogue survey, dialogues, and a post-dialogue survey.

Stage 1. The pre-dialogue survey measured participants' demographics including race/ethnicity, gender, age, income, education, and political ideology. To assess their efficacy in chatbot-related experiences, we drew from existing measurements of consumer experience with technology (22) and public responses to AI (32). Participants were then divided into two groups to discuss one of the following crucial issues: climate change or BLM. Before the conversation, we measured participants' attitudes toward the two issues. For participants in the climate change group, we asked them questions like whether they perceive climate change as a real phenomenon; one that is due to human activities and has negative impacts. For participants in the BLM group, we asked them questions such as whether they support the movement and if they perceive it as necessary. The measurements achieved internal consistency with Cronbach's alphas ranging from 0.86 to 0.88. These attitudinal questions were asked again in the post-dialogue survey and were used to calculate participants' knowledge gains on the two issues.

Stage 2. Participants were then directed to our UI webpage (see SI Appendix A1.1) to have dialogues with GPT-3 on their assigned topic. Each participant was required to have anywhere between six to 12 rounds of conversation with the chatbot. The whole dialogue was organic. We did not manipulate the chatbot to fit either of the topics, hence the dialogue topics were initiated by participants (i.e., participants need to ask questions about climate change or express ideas about BLM first).

Stage 3. The post-dialogue survey assessed participants' evaluations of their user experience with the chatbot. Five sets of questions were provided for participants to evaluate their user experiences (1) ratings of the chatbot, (2) satisfaction with the dialogue, (3) learning experience with the chatbot, and their intention to (4) continue the chat or (5) recommend the chatbot to others. These evaluation questions were measured on a 5-point Likert scale that asks the participants to indicate to what extent they agree with the statements. Cronbach's alpha suggests high internal consistency in these measurements, ranging from 0.88 to 0.94. We further had our participants respond to open-ended questions, writing what they expected to hear from GPT-3, but GPT-3 failed to provide.

We conducted pilot tests in November 2021 to refine our survey instruments and our participant-GPT-3 chat UI design. Then we entered the real launch, during which we recruited 4,240 anonymous participants from the crowdsourcing platform Amazon Mechanical Turk from December 2021 to February 2022. To ensure data quality, those who failed the attention check questions or completely stayed off-topic during dialogues were marked as invalid workers and were excluded from the analyses. In the end, we had 3,290 valid participants, resulting in 26,211 rounds of dialogues with GPT-3. Table S1 in SI Appendix 2.2 presents the demographic distribution of our valid participants and their pre-chat attitudes toward climate change and BLM.

We re-coded participants' race and ethnicity, primary language, education level, and prior attitudes on climate change and BLM into majority *vs.* minority groups (i.e., a binary variable). 21.49% of participants self-identified as non-white in our recorded sample and were recoded as the minority group for the race and ethnicity variable. The rest 80% of participants who self-identified as white were recoded as the majority group. Further, 8.81% of our participants' first language is not English. There is also a significant education attainment divide in our collected sample. Approximately 82.22% of our sample have at least acquired a bachelor's degree (i.e., the majority group). We recoded participants with lower than bachelor's degrees as education minorities. We recoded participants' opinions toward climate change into minorities *vs*. majorities, using the 1st quarter of the average attitude score as the threshold, where a higher score means more supportive of climate change facts. Similar dividing criteria were applied to participants in the BLM topic to divide them into the opinion minority *vs.* opinion majority group.

*Analysis Method*

We conducted OLS regressions to analyze the relationship between users' demographics, attitudes towards the issue, and their user experience after the chat (RQ1), as well as associations between GPT-3's conversational features and the resulting user experiences (RQ3). Our data distribution exhibited a highly bi-modal style, with most user experience items gathered around either the lower quantile (i.e., extremely negative) or the higher quantile (i.e., extremely positive). Thus, we also conducted quantile regressions as a robust check on our OLS regression results. More information about this robust check can be found in SI Appendix A3.3.

To analyze dialogues (RQ2), we conducted topic modeling and linguistic analyses. Specifically, we used structural topic modeling (STM) (33), which produces the most prevalent topics and associated keywords from a set of documents based on the Latent Dirichlet Allocation (LDA) approach, while also considering prior differences in users' demographics and attitudes. We generated the most prevalent ten topics for every round of participant-chatbot conversation in the climate change and BLM dialogues respectively, with participants' demographic divides and the order of rounds in the dialogue as covariates. To further examine conversational styles, we also used the Linguistic Inquiry and Word Count (LIWC) software (34), which provides word counts of positive emotion and negative emotion, analytic words showing logical thinking, clout words showing confidence, and authentic words showing genuineness.

We performed stance detection analyses on the dialogues between GPT-3 and participants to gain insights into why participants, on average, developed more supportive attitudes toward human-caused climate change and BLM. Utilizing Recurrent Neural Networks (RNN), we classified the prompts into dichotomous variables of "supporting" or "not supporting" of climate change/BLM. We achieved a high level of accuracy in detecting stance in both climate change (F1 score = 0.74) and BLM (F1 score = 0.78). Further details about stance detection are in SI Appendix A3.5.

**References**

1. E. Hosseini-Asl, B. McCann, C.-S. Wu, S. Yavuz, R. Socher, A simple language model for task-oriented dialogue. *Advances in Neural Information Processing Systems* **33**, 20179-20191 (2020).
2. N. Mirchi *et al.*, The Virtual Operative Assistant: An explainable artificial intelligence tool for simulation-based training in surgery and medicine. *PloS one* **15**, e0229596 (2020).
3. S. J. Cachumba, P. A. Briceño, V. H. Andaluz, G. Erazo (2019) Autonomous driver assistant for collision prevention. in *Proceedings of the 2019 11th International Conference on Education Technology and Computers*, pp 327-332.
4. J. Zhang, Y. J. Oh, P. Lange, Z. Yu, Y. Fukuoka, Artificial intelligence chatbot behavior change model for designing artificial intelligence chatbots to promote physical activity and a healthy diet. *J. Med. Internet Res.* **22**, e22845 (2020).
5. M. D. McCradden, S. Joshi, M. Mazwi, J. A. Anderson, Ethical limitations of algorithmic fairness solutions in health care machine learning. *The Lancet Digital Health* **2**, e221-e223 (2020).
6. L. Weidinger *et al.* (2022) Taxonomy of risks posed by language models. in *2022 ACM Conference on Fairness, Accountability, and Transparency*, pp 214-229.
7. A. Koenecke *et al.*, Racial disparities in automated speech recognition. *Proceedings of the National Academy of Sciences* **117**, 7684-7689 (2020).
8. The Sentinel (2022) Google wants you to chat with its Artificial Intelligence chatbot at your own risk.
9. K. A. Raaflaub, *Equalities and inequalities in Athenian democracy* (Princeton University Press Princeton, NJ, 1996).
10. W. Von Leyden, *Aristotle on equality and justice: His political argument* (Springer, 1985).
11. J. J. Mansbridge, *Beyond adversary democracy* (University of Chicago Press, 1983).
12. A. Gutmann, *Liberal equality* (CUP Archive, 1980).
13. K. Chen, J. Jeon, Y. Zhou, A critical appraisal of diversity in digital knowledge production: Segregated inclusion on YouTube. *New Media & Society*, 14614448211034846 (2021).
14. J. Fishkin, *When the people speak: Deliberative democracy and public consultation* (Oup Oxford, 2009).
15. A. Lupia, A. Norton, Inequality is always in the room: language & power in deliberative democracy. *Daedalus* **146**, 64-76 (2017).
16. I. Freiling, N. M. Krause, D. A. Scheufele, K. Chen, The science of open (communication) science: Toward an evidence-driven understanding of quality criteria in communication research. *Journal of Communication* **71**, 686-714 (2021).
17. S. Jasanoff, *The ethics of invention: Technology and the human future* (WW Norton & Company, 2016).
18. R. Owen, J. R. Bessant, M. Heintz, *Responsible innovation: managing the responsible emergence of science and innovation in society* (John Wiley & Sons, 2013).
19. *For a list of recent publications on fairness in NLP, please see: https://github.com/uclanlp/awesome-fairness-papers#dialogue-generation*
20. E. Noelle-Neumann, The spiral of silence: A theory of public opinion. *Journal of Communication* **24**, 43-51 (1974).
21. Q. V. Liao *et al.* (2018) All work and no play? Conversations with a Question-and-Answer Chatbot in the Wild. in *Proceedings of the 2018 CHI Conference on Human Factors in Computing Systems*, pp 1-13.
22. A. Venkatesh *et al.*, On evaluating and comparing conversational agents. *arXiv preprint arXiv:1801.03625* **4**, 60-68 (2018).
23. W. Liao, J. Zhang, Y. J. Oh, N. A. Palomares, Linguistic accommodation enhances compliance to charity donation: The role of interpersonal communication processes in mediated compliance-gaining conversations. *Journal of Computer-Mediated Communication* **26**, 167-185 (2021).
24. A. B. Wassink, C. Gansen, I. Bartholomew, Uneven success: automatic speech recognition and ethnicity-related dialects. *Speech Communication* **140**, 50-70 (2022).
25. J. Steiner, *The foundations of deliberative democracy: Empirical research and normative implications* (Cambridge University Press, 2012).

26. S. J. O'Neill, M. Boykoff, Climate denier, skeptic, or contrarian? *Proceedings of the National Academy of Sciences* **107**, E151-E151 (2010).
27. R. R. Mourão, D. K. Brown, Black Lives Matter coverage: How protest news frames and attitudinal change affect social media engagement. *Digit. Journal.* **10**, 626-646 (2022).
28. A. Birhane *et al.* (2022) The forgotten margins of AI ethics. in *2022 ACM Conference on Fairness, Accountability, and Transparency*, pp 948-958.
29. C. Yan, J. P. Dillard, F. Shen, Emotion, motivation, and the persuasive effects of message framing. *Journal of Communication* **62**, 682-700 (2012).
30. L. Festinger, Cognitive dissonance. *Scientific American* **207**, 93-106 (1962).
31. E. Harmon-Jones, J. W. Brehm, J. Greenberg, L. Simon, D. E. Nelson, Evidence that the production of aversive consequences is not necessary to create cognitive dissonance. *Journal of personality and social psychology* **70**, 5 (1996).
32. S. Cave, K. Coughlan, K. Dihal (2019) " Scary Robots" Examining Public Responses to AI. in *Proceedings of the 2019 AAAI/ACM Conference on AI, Ethics, and Society*, pp 331-337.
33. M. E. Roberts, B. M. Stewart, D. Tingley, Stm: An R package for structural topic models. *Journal of Statistical Software* **91**, 1-40 (2019).
34. R. L. Boyd, A. Ashokkumar, S. Seraj, J. W. Pennebaker, The development and psychometric properties of LIWC-22. *Austin, TX: University of Texas at Austin* (2022).

**Figures**

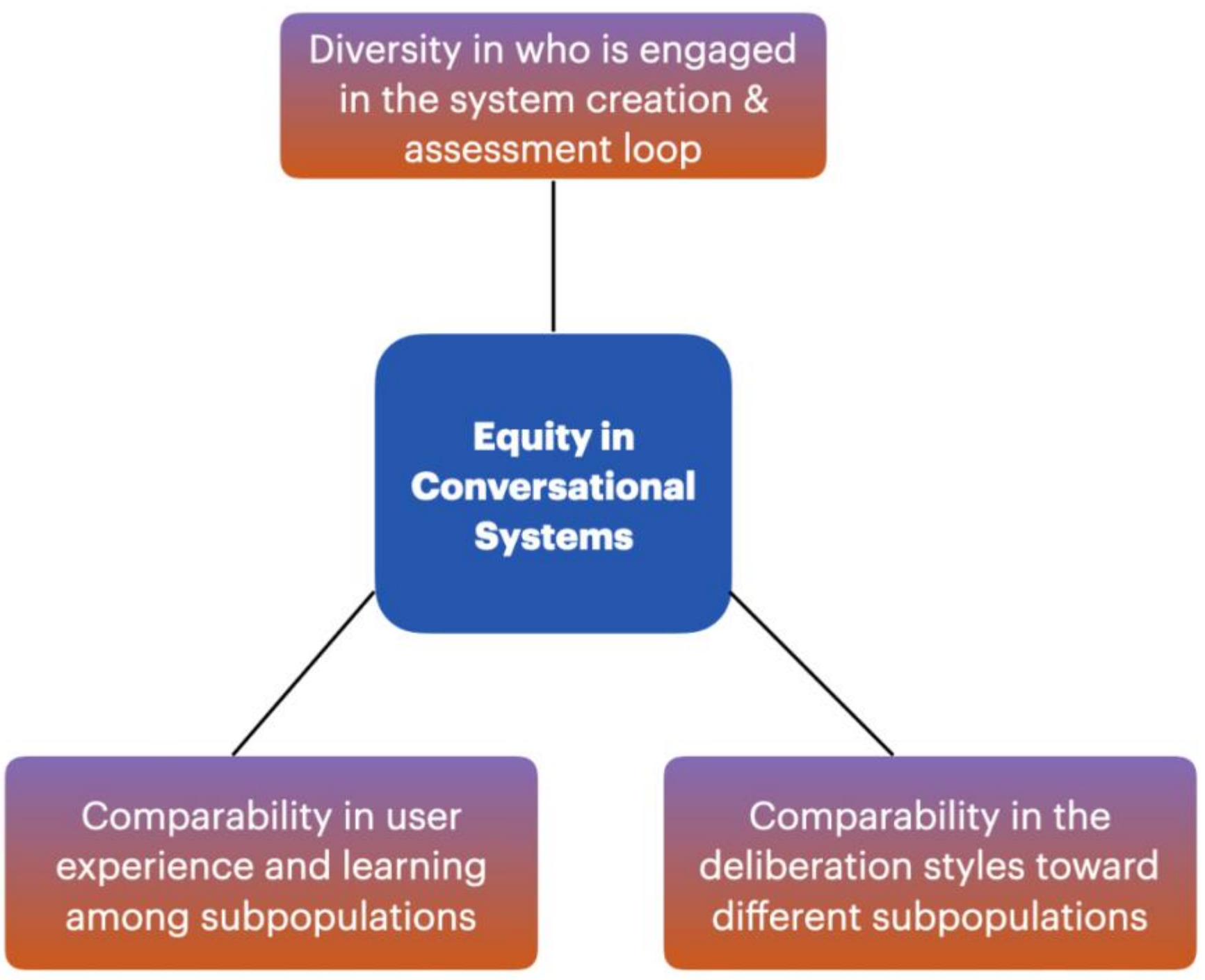

**Figure 1.** A Framework for Assessing Equity in Conversational AI

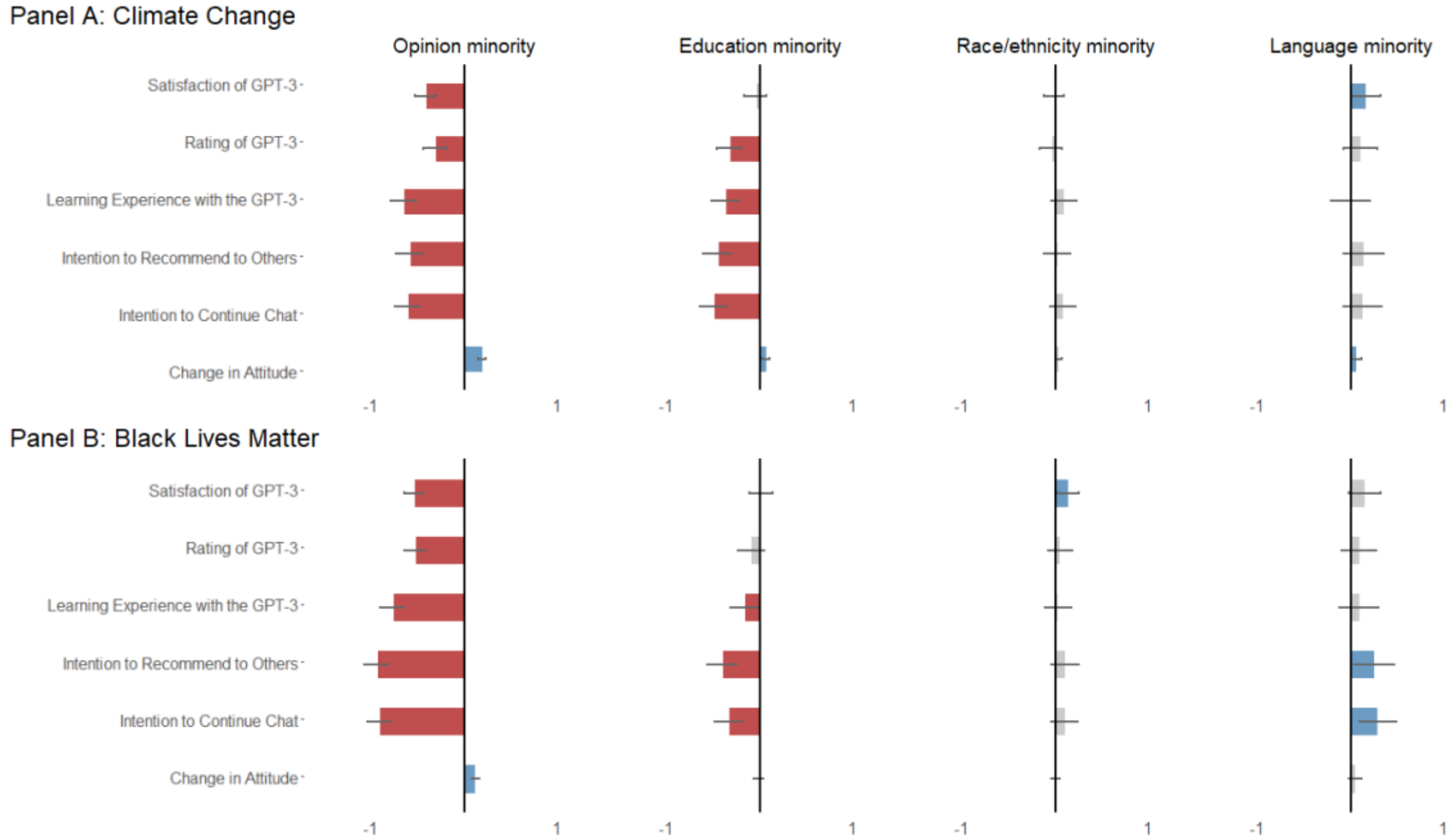

**Figure 2.** Associations Between User's Demographic Attributes and their User Experience After Chatting with GPT-3 on Climate Change (Panel A) and on BLM (Panel B)

Note: Each plot in a panel represents an aggregated summary for one user demographics (e.g., minority status in opinion) and its performance in the six different linear regression models for six user experience variables (e.g., satisfaction with the chatbot). The bars represent the coefficient values, while the error bars represent the confidence intervals at 95%. Statistically significant negative coefficients are marked in red, significant positive coefficients are marked in blue, and non-significant ones are marked in grey. Each plot presents part of the full regression models. In the full regression models, we controlled for other demographic variables including participants' age, income level, previous experience and knowledge with using chatbots, as well as each participant's language styles such as the word count of their average input, use of positive and negative emotion words, use of analytical words, clout, and authentic expressions in conversations. For the full OLS regression tables for each plot in Panel A and Panel B, please check Table S2, S3, and S4 in SI Appendix A3.3.

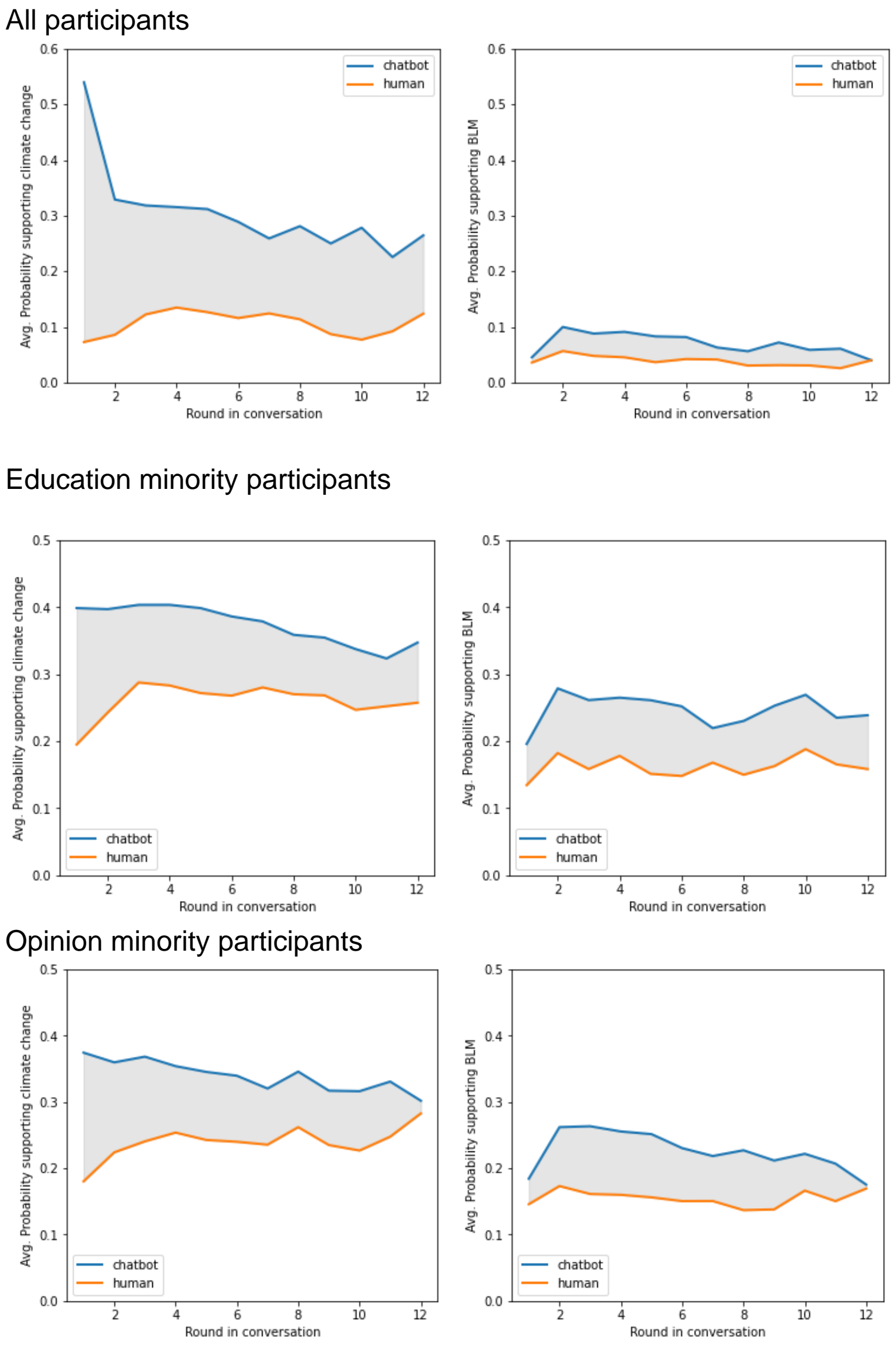

**Figure 3.** How Issue Stance Evolved throughout Conversations for Participants and GPT-3

**Panel A: Climate Change**

**(1). STM Analyses**

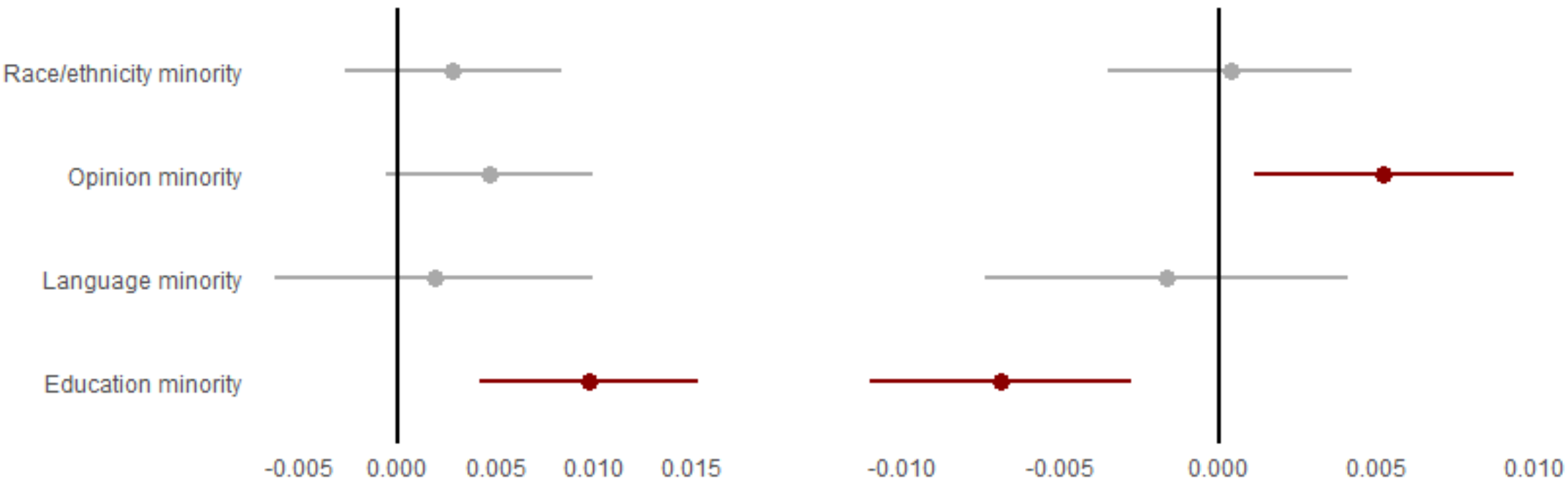

**(2). LIWC Analyses**

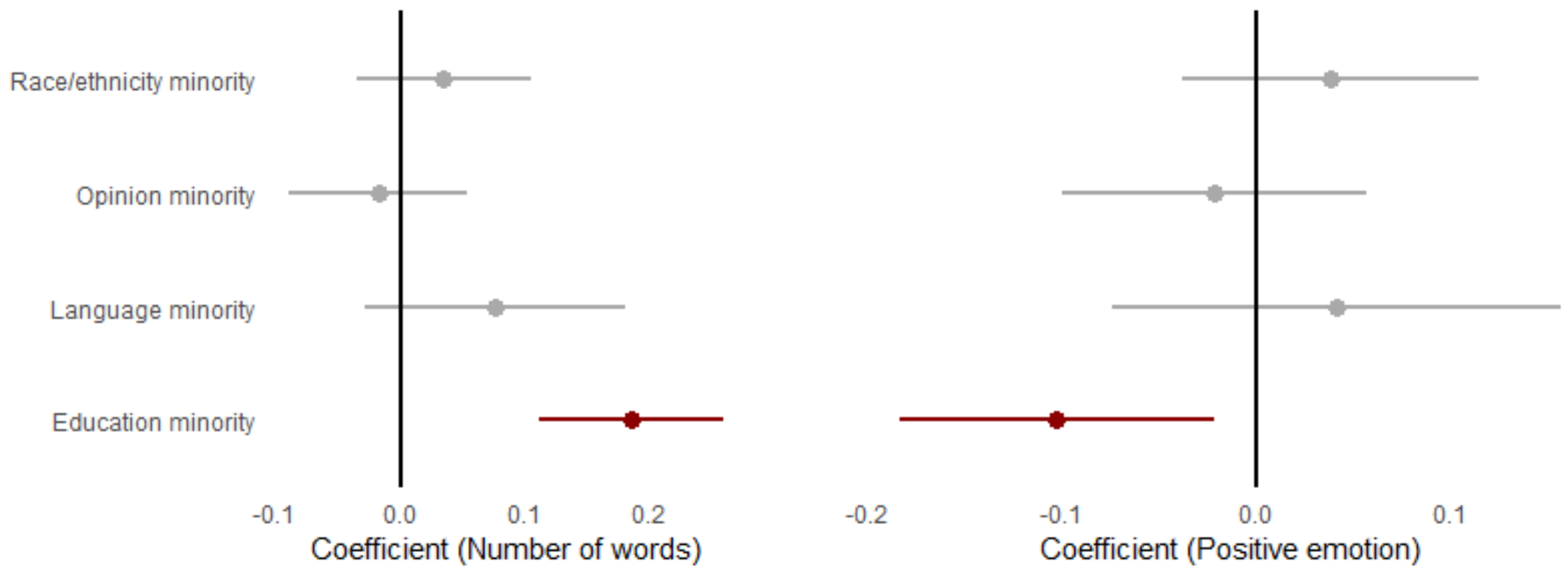

**Panel B: BLM**

**(1). STM Analyses**

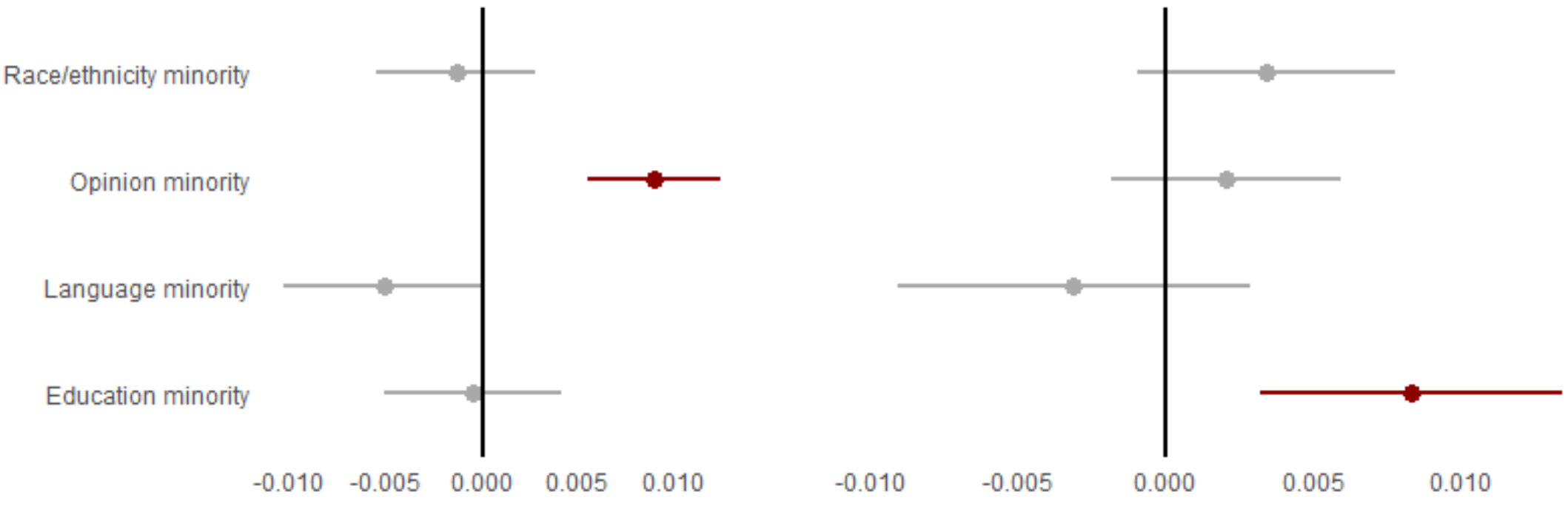

**(2). LIWC Analyses**

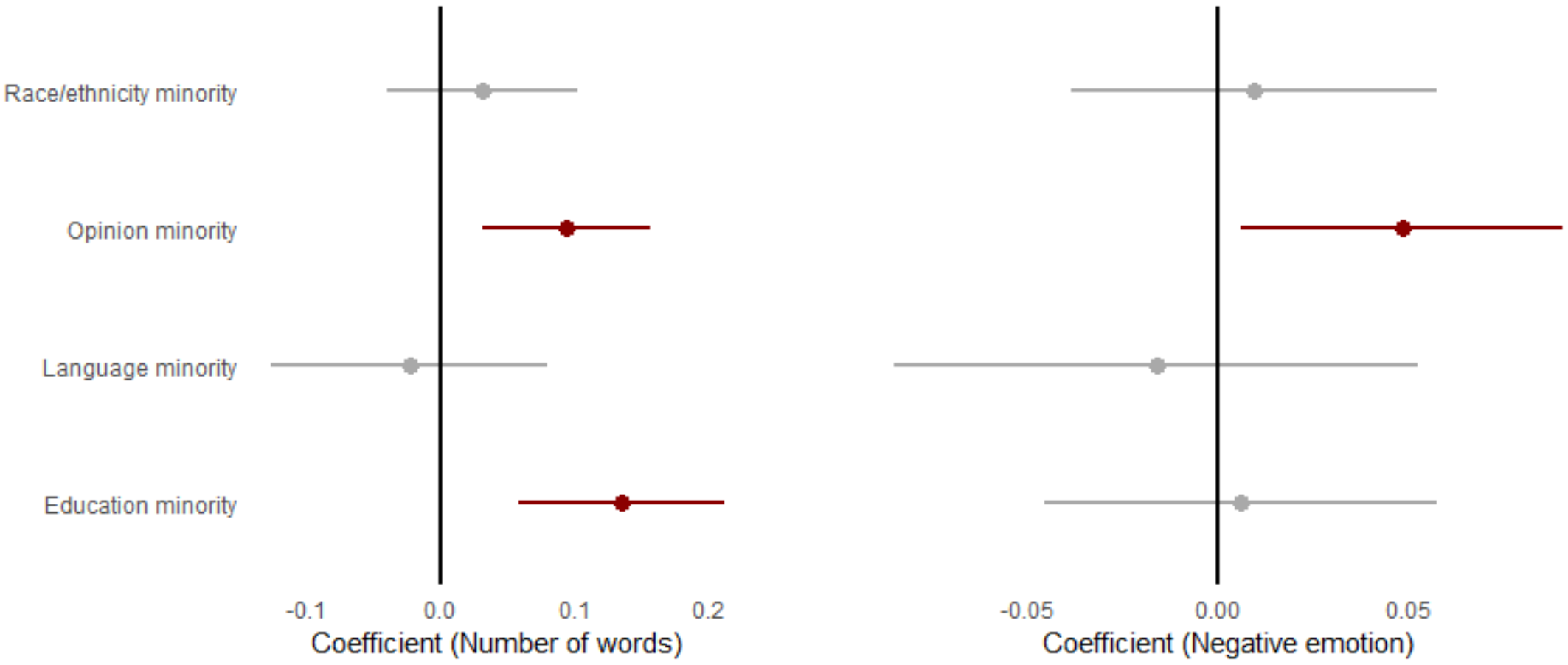

**Figure 4.** Effects of Participants' Demographics on GPT-3's Response Content and Style During the Climate Change Conversation (Panel A) and BLM Conversation (Panel B). Within each Panel, we present the results from: (1) STM analyses. The dots represent the average effect of an IV on the prevalence of a topic. 95% Confidence Intervals are included. (2) LIWC Language features such as the Number of Words and the Use of Positive/Negative Emotions.

Note: Please refer to SI Appendix 3.4 Table S10 for an examination of all the STM topics and their interpretations for the responses GPT-3 gave during the climate change and BLM discussions. Table S11 in SI Appendix 3.4 provides statistical reports about the associations between our covariates (e.g., demographic variables) and a topic's prevalence. For a full analysis of the LIWC results presented in this figure, please see Table S5 in SI for the climate change issue and Table S6 in SI for the BLM issue.

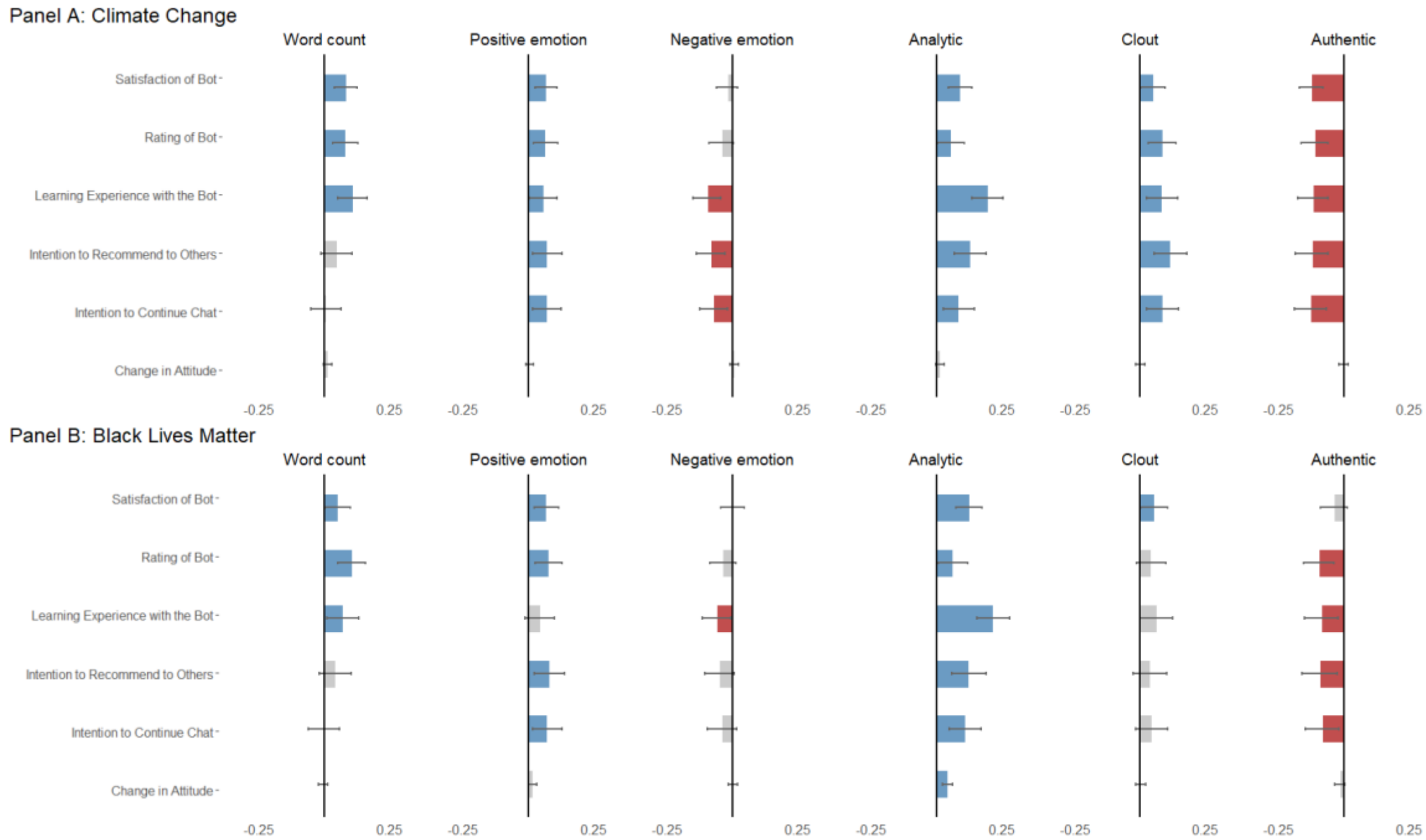

**Figure 5.** Associations between GPT-3's Conversational Styles and User Experience

Note: To improve the scaling and interpretability of the word count and emotion features in the regression model, we applied a log transformation (log(x+1)) to these variables. In this visualization, we standardized all numeric variables using z-scores to facilitate the comparison of the coefficients. Statistically significant coefficients are highlighted in blue (positive coefficients) or red (negative coefficients). Full regression tables (variables are non-standardized) can be found in SI Table S7, S8, and S9.